\pdfoutput=1

\documentclass[11pt]{article}
\usepackage{acl}
\usepackage{times}
\usepackage{latexsym}

\usepackage{lipsum}

\usepackage{microtype}

\usepackage{amsmath}
\usepackage{makecell}
\usepackage{graphicx}
\usepackage{booktabs}
\usepackage{multirow}
\usepackage{siunitx}
\usepackage{enumitem}
\usepackage{colortbl}
\usepackage{xcolor}
\usepackage{url}
\usepackage{amssymb}
\usepackage{amsthm}
\usepackage{mathtools}
\usepackage{subcaption}
\usepackage{makecell}
\usepackage{xspace}
\usepackage[ruled,vlined]{algorithm2e}
\usepackage[warn]{textcomp}
\usepackage{ marvosym }
\usepackage{stfloats}   
\usepackage{dcolumn}

\newcommand{\com}[1]{}

\newcolumntype{d}[1]{D{.}{.}{#1}}

\makeatletter
\def\blfootnote{\xdef\@thefnmark{}\@footnotetext}
\makeatother

\title{End-to-End Speech Translation for Code Switched Speech}

\author{Orion Weller$^{1}$*, Matthias Sperber$^{2}$,  Telmo Pires$^{2}$, \\ \textbf{Hendra Setiawan$^{2}$, Christian Gollan$^{2}$, Dominic Telaar$^{2}$, Matthias Paulik$^{2}$} \\ 
$^1$Johns Hopkins University \\
$^2$Apple\\
\normalsize{\texttt{oweller@cs.jhu.edu,sperber@apple.com}}
}

\date{}

\begin{document}
\maketitle
\begin{abstract}
Code switching (CS) refers to the phenomenon of interchangeably using words and phrases from different languages. CS can pose significant accuracy challenges to NLP, due to the often monolingual nature of the underlying systems. In this work, we focus on CS in the context of English/Spanish conversations for the task of speech translation (ST), generating and evaluating both transcript and translation. To evaluate model performance on this task, we create a novel ST corpus derived from existing public data sets.\footnote{We make instructions and extra data needed to construct our CS data set available at \url{https://github.com/apple/ml-code-switched-speech-translation}} We explore various ST architectures across two dimensions: cascaded (transcribe then translate) vs end-to-end (jointly transcribe and translate) and unidirectional (source $\rightarrow$ target) vs bidirectional (source $\leftrightarrow$ target). We show that our ST architectures, and especially our bidirectional end-to-end architecture, perform well on CS speech, even when no CS training data is used.

\end{abstract}

\section{Introduction}

Over half of the world's population is estimated to be bilingual.
\footnote{BBC: \url{https://bbc.in/3jgwzZ2}} \blfootnote{\text{*} Work done as an intern.}Those that know multiple languages are prone to code switch, i.e., to interchangeably use words and phrases from two (or more) languages in situations such as casual dialog, while traveling abroad, or simply to use a word they find more fitting \cite{scotton1977bilingual,heredia2001bilingual}. In CS, the base language is referred to as the \textit{matrix} language while the contributing language is called the \textit{embedded} language \cite{myers1995social}, where speakers often use the matrix language the majority of the time.

\begin{figure}[t!]
    \centering
    \includegraphics[trim=5 10 40 0,clip,width=0.45\textwidth]{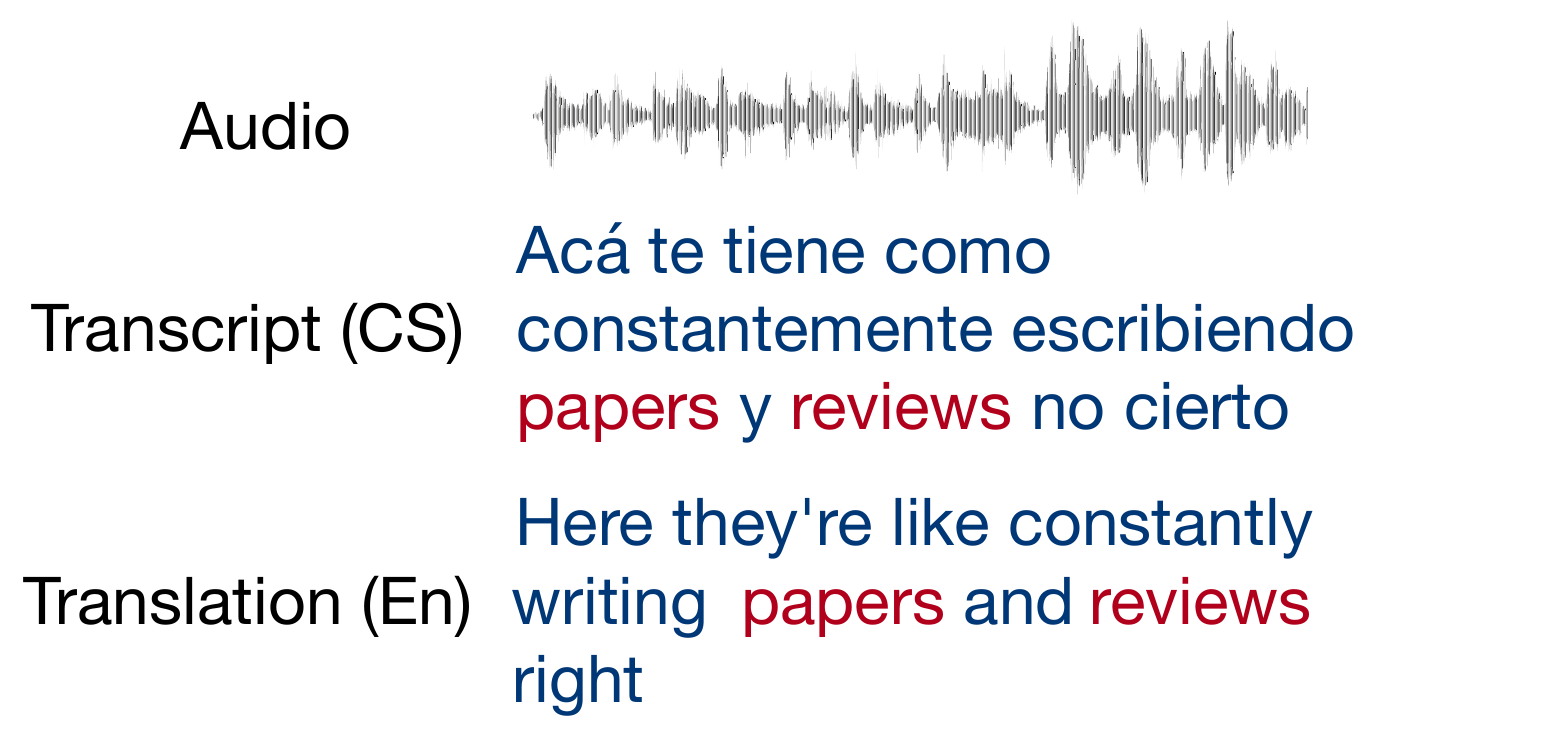}
    \caption{An example instance of the joint speech recognition and translation task for code-switching (CS). Red indicates English words in the transcript and their corresponding words in the translation, whereas blue indicates Spanish words in the transcript and their corresponding translation.}
    \label{fig:task}
\end{figure}

Code switched language is challenging to both automatic speech recognition (ASR) and machine translation (MT) - and therefore also to the composite task of speech translation (ST). While a rich amount of prior works exist on CS in the context of ASR \cite{lyu2006speech,ahmed2012automatic,vu2012first,10.1162/tacl_a_00065,yue2019end} and MT \cite{sinha2005machine,Winata2021AreMM,zhang2021rnn,yang2020csp}, there is little prior work in the context of ST.

The aforementioned challenges to ASR, MT and ST arise largely due to the lack of CS data as well as the often monolingual nature of ASR systems, and of encoders of MT and ST systems. The lack of CS data is often addressed via synthetic data, e.g. as seen in \citet{xu2021can,Nakayama2019RecognitionAT}. Instead, in this work we derive two novel natural CS datasets from existing public corpora.
CS is also difficult for modeling due to its mixed multilingual nature. In order to support multiple languages on the utterance level, automatic language identification (LID) is often performed before applying monolingual systems on a per utterance basis. However, this does not address within-utterance CS, where embedded foreign words and phrases result in recognition errors for monolingual ASR systems, making multilingual models an attractive alternative. Furthermore, CS increases speech recognition errors, significantly increasing the problem of error propagation \cite{ruiz2014assessing} in cascaded ST systems, where MT is then performed on the erroneous ASR output. Thus, multilingual end-to-end (E2E) ST systems may be especially appropriate to tackle CS speech.

As both the transcript and translation are important in many CS ST use cases, we focus on the joint transcription and translation ST setting \cite{anastasopoulos2018tied,weller2021streaming}, extending it to CS data. We follow the methodology of these previous works and focus on the triangle E2E ST model to jointly generate both a transcript of the CS utterance and a translation of that utterance into text containing only one language (c.f. Figure~\ref{fig:task} for an illustration). We perform a comparison along two axes: (1) comparing this E2E model to the standard cascaded ST systems, and (2) exploring the difference between bilingual systems and primarily monolingual systems gated by utterance-level LID. Following recent work that has shown the effectiveness of pre-trained models for ST \cite{li2020multilingual,gallego2021upc}, we use Wav2Vec 2.0 \cite{baevski2020wav2vec} as our encoder model and the multilingual mBART 50-50 \cite{tang2020multilingual} as our decoder model.

We also make several modeling contributions in order to use these pre-trained models for joint transcription and translation. For the E2E ST model, we extend \citet{li2020multilingual} to adapt the mBART decoder to jointly produce both transcription and translation. Furthermore, we introduce a triangle E2E ST model with a shared bilingual decoder and show that this improves transcription and translation accuracy. Our model analysis shows a surprising amount of robustness to CS speech, with the amount (or proportion) of CS words in a sentence not affecting model accuracy. Overall, we observe strong accuracy scores (WER, BLEU) on the CS task, both without CS training data and in the low-resource setting. We believe this opens the door to new and exciting progress in this area.\\

\section{Related Work}
\label{sec:related}

Code-switching in NLP has seen a rise of interest in recent years, including a dedicated workshop starting in 2014 \cite{diab2014proceedings} and still ongoing \cite{calcs-2021-approaches}. CS in machine translation also has a long history \cite{le1990different,climent2003bilingual,sinha2005machine,10.1162/tacl_a_00065,elmadany2021investigating,xu2021can}, but has seen a rise of interest with the advent of large multilingual models such as mBART \cite{liu2020multilingual} or mT5 \cite{xue2020mt5,gautam2021comet,jawahar2021exploring}. Due to the lack of available CS data and the ease of single-word translation, most of these recent related MT works have synthetically created CS data for either training or testing by translating one or more of the words in a sentence \cite{Song2019CodeSwitchingFE,Nakayama2019RecognitionAT,xu2021can,yang2020csp}. We differ from those works by using naturally occurring CS data (Section~\ref{sec:taskanddata}) which models the real-world CS distribution rather than arbitrary language mixing.

For spoken input, as present in ASR and ST, synthetically creating realistic CS data is more challenging than it is for MT. However, dedicated ASR corpora that contain natural CS exist, including the Bangor Miami \cite{deuchar20145}, SEAME \cite{zeng2018end}, and the recent large-scale ASRU 2019 task \cite{shi2020asru}. These corpora generally do not contain translations of the ASR annotations, since they were designed for the ASR task only. However, there exist two exceptions, which we leverage to derive our ST CS data set, described in Section~\ref{sec:taskanddata}. 

There also exists a wide range of prior modeling work on CS in ASR models, for a variety of strategies \cite{lyu2006speech,ahmed2012automatic,Seki2018AnEL,luo2018towards,lu2020bi,du2021data,zhang2021rnn}. However, the recently introduced large multilingual models for speech, such as Wav2Vec, Wav2Vec 2.0, \citet{schneider2019wav2vec,baevski2020wav2vec} and HuBERT \cite{hsu2021hubert}, are still underexplored with regards to their CS performance. 

Handling mixed languages also requires understanding what languages are being spoken. Systems that support mixed language input therefore require some form of automatic LID -- either as an explicit component on the utterance  \cite{mabokela2014modeling,xu2021can} or word-level \cite{Lyu2008LanguageIO,Nakayama2019RecognitionAT}, or implicitly learned by the underlying model(s) via a multi-task learning setup \cite{lyu2008language,watanabe2017language,hou2020large}. In our work, we leverage both, exploring utterance-level LID components as well as implicit learning of utterance and word level LID.

In both MT and ASR, prior publications have also included the study of intra-word mixing of languages \cite{yilmaz2018building,mager2019subword}, a phenomenon we do not explore in our work.

Finally, our work builds off of advances made by \citet{gallego2021upc,li2020multilingual} that show that combining large multilingual speech and text models provide consistent improvements. We differ however, by exploring ST in the novel CS setting. 

\section{Task Description \& Data Used}
\label{sec:taskanddata}

\subsection{Task Description}
We investigate systems suitable for bilingual English/Spanish conversational scenarios where some of the English and Spanish utterances may include some amount of words and phrases of the respective other language. That is, we are focusing on ST systems that can automatically and seamlessly handle utterances that are either purely English, purely Spanish, English with some Spanish words/phrases embedded or Spanish with some English words/phrases embedded. For transcription, we aim for models to generate the exact mixed-language transcript with each word written in its original spoken language. For translation, we aim to generate purely monolingual translations. See Figure~\ref{fig:task} for an example. The experiments and results presented in this paper focus on translating into monolingual English only due to data availability, although we expect similar results for Spanish translations, due the bidirectional model training on standard ST data (Appendix~\ref{app:training_test}). We will leave it to future work to more closely examine translation into Spanish -- or even a third language not present in the original utterance.

It must be noted that word-level language categorization is sometimes ambiguous. A word in one language may also be considered part of a different language. That is for example true for \textit{loan words}  \cite{baugh1935chronology}, e.g., \textit{e-mail} in many non-English languages such as German. This issue can be further complicated by attempting to categorize what language named entities fall under: is a Spanish speaker saying \textit{Joe Biden} or \textit{New York} code-switching? Although we acknowledge the complexity of separating words between languages, our work, following previous work \cite{Modipa2013ImplicationsOS,Nakayama2018SpeechCF}, uses data annotated by crowd-sourced workers, counting any sentence annotated as having a least one foreign word as being CS. This approach also makes intuitive sense for speech, as the CS words (classified as foreign) will have phonemes that will align more with the embedded language, while the non-CS phonemes will align more with the matrix language.

\definecolor{paperred}{RGB}{157,3,40}
\definecolor{paperblue}{RGB}{37,53,109}
\begin{table*}[t!]
    \centering
    \begin{tabular}{lp{9cm}p{4cm}}
      Dataset & Raw Transcript & Clean Transcript \\
    \toprule
      Fisher & \textcolor{paperblue}{un} \textless foreign lang="English"\textgreater\
  \textcolor{paperred}{show} \textless$\backslash$foreign\textgreater, \textcolor{paperblue}{a mi me gusta ver mucho estos} \textless foreign lang="English"\textgreater\ \textcolor{paperred}{shows} \textless$\backslash$foreign\textgreater\ \textcolor{paperblue}{de la medicina forense} & \textcolor{paperblue}{un} \textcolor{paperred}{show}, \textcolor{paperblue}{a mi me gusta ver mucho estos} \textcolor{paperred}{shows} \textcolor{paperblue}{de la medicina forense} \\
  \hline
  Miami & \textcolor{paperblue}{hay una} [/] \textcolor{paperblue}{una que dice} (.) \textcolor{paperred}{it's}@s:eng \textcolor{paperred}{five}@s:eng \textcolor{paperred}{o'clock}@s:eng \textcolor{paperred}{somewhere}@s:eng &  \textcolor{paperblue}{hay una una que dice} \textcolor{paperred}{it's five o'clock somewhere} \\

      \bottomrule
    \end{tabular}
    \caption{Examples of the raw and clean data for Miami and Fisher. Text in red indicates English text while blue text indicates Spanish. The Miami dataset uses the CHAT annotation format \cite{macwhinney1990child}.}
    \label{tab:raw}
\end{table*}

\subsection{Code-Switched Speech Datasets}

We use the Fisher \cite{cieri2004fisher} and Bangor Miami\footnote{Online audio files can be found at \url{https://biling.talkbank.org/access/Bangor/Miami.html}} \cite{deuchar20145} corpora for CS data, as they are the only publicly available corpora we are aware of that contains both annotated CS ASR transcripts, as well as translations of those transcripts (Table~\ref{tab:raw}). Although these corpora contain the translations, to our knowledge they have not been used to study CS translation before.

The Miami corpus was collected for linguistic code-switching analysis and gathered from recorded conversations between bilingual English/Spanish speakers in casual settings, primarily in Miami, Florida. These conversations include a high proportion of naturally occurring CS speech. However, in order to collect these naturally occurring conversations, the participants were recorded throughout their day using a small digital recorder worn on belts and lapels. Due to this, the Miami audio contains lower audio quality and much noiser background conditions than standard ASR datasets. 

The Fisher dataset was collected for ASR and was gathered by pairing sets of Spanish speakers, located in the U.S. and Canada, to each other through phone calls. Although the Fisher dataset is not a CS focused dataset, we found that it contains a large amount of (annotated) CS utterances, due to the speakers being situated in English-speaking contexts.  The recording method (phone recordings in 2004) makes this a noisy ASR dataset, although significantly less so than Miami.

\begin{table}[t]
\small
\centering
\setlength\extrarowheight{2pt}
\begin{tabular}{llrrr}
\toprule
Dataset & Split & Type & Hours & Instances \\ \midrule
\multirow{3}{*}{Miami} & Train & Mono & 3.60 & 6,489 \\
\cline{2-5}
 & \multirow{2}{*}{Test} & CS & 2.82 & 3,296 \\
&   & Mono & 3.61 & 6,490 \\

\midrule
\multirow{5}{*}{Fisher} & \multirow{2}{*}{Train} & CS & 13.28 & 7,398 \\
&  & Mono & 157.3 & 130,600   \\
\cline{2-5}
& Dev & CS & 1.45 & 821 \\
\cline{2-5}
& \multirow{2}{*}{Test} & CS & 1.63 & 986 \\
& & Mono & 12.15 & 10,595 \\
\bottomrule
\end{tabular}
\caption{Dataset Statistics. CS stands for Code-Switched and Mono for Monolingual.}
\label{tab:dataset_stats}
\end{table}

\begin{figure*}[htb]
    \centering
    \includegraphics[trim=0 0 0 0,clip,width=0.48\textwidth]{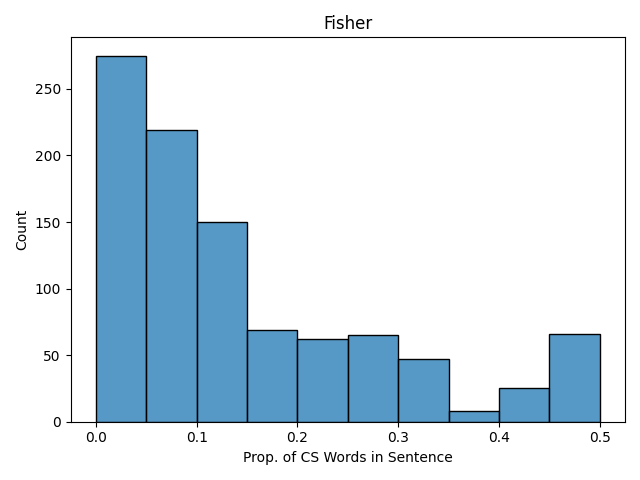}
    \includegraphics[trim=0 0 0 0,clip,width=0.48\textwidth]{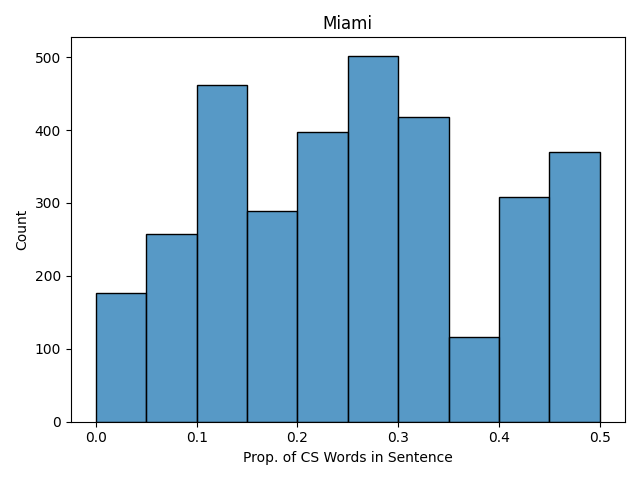}
    \caption{Histogram of the proportions of code-switched words in a sentence for the CS test sets (Fisher on the left, Miami on the right). For example, 0.2 means that 20\% of the words in the sentence are CS.}
    \label{fig:cs_distributions}
\end{figure*}

To prepare the data for the joint ST CS task, we separate the data with CS utterances (utterances that contain at least one word annotated as CS) from those with none, creating a CS set and a monolingual set for each dataset. We note that for the Miami dataset the monolingual split contains both English-only and Spanish-only monolingual audio. As the Miami corpus was also annotated with both ambiguous and unambiguous code-switching, we only include utterances in the CS set if the annotations were tagged as unambiguously code-switched (i.e. excluding words such as \textit{ok}, \textit{aha}, and named entities). The Fisher CS dataset consists of majority (matrix\footnote{For simplicity, we use the terms majority/matrix language and minority/embedded language interchangeably.}) Spanish 77\% of the time, English-majority 17\%, and 6\% evenly split between English/Spanish. For the Miami CS dataset the languages are more evenly distributed, with 51\% majority-Spanish, 35\% majority-English, and 9\% evenly split.\footnote{To make these CS datasets reproducible for the broader ST community, we provide a file with instructions for gathering the data (as Fisher is part of the LDC library) as well as files containing a mapping between the original dataset indices to the CS data splits.}

The Fisher data consists of three evaluation sets (Dev/Dev2/Test) that together contain approximately a thousand instances of CS with corresponding translations in monolingual English. We combine them into a Fisher CS \textit{Test} set. The Fisher dataset also contains a large amount of CS utterances in the training set (appx. 8k or 15 hrs) which we use as fine-tuning (90\%) and validation data (10\%). As the Miami dataset contains no splits, we use all CS data for the test set and split the monolingual data into even train/test sets.  We include basic summary statistics in Table \ref{tab:dataset_stats}. Note that when compared to standard ST datasets, these CS ST datasets would be considered low-resource settings.

In Figure~\ref{fig:cs_distributions}, we see the proportion of CS words in a sentence for the CS test sets. We note that there are no sentences with more than 50\% of the words CS since the minority language cannot be more than 50\% by definition. For instances that are exactly 50\% code switched their language identification was chosen by randomly selecting either English or Spanish. We see that for the Fisher dataset there are more sentences with less than 15\% CS with a small uptick around 50\%.  For Miami it is more uniform, with a large amount of sentences being approximately 25\% CS.

To prepare our models for Spanish-English CS, we use the CoVoST \cite{wang2020covost,wang2020covost2} and MuST-C \cite{cattoni2021must} datasets for standard ST training, as CoVoST contains only Es$\xrightarrow{}$En and MuST-C contains only En$\xrightarrow{}$Es.  Although high scores on these datasets are not our primary target, we note that our scores come close to or improve the state of the art (SoTA) on these tasks (see Appendix~\ref{app:training}, Table \ref{tab:training}) albeit with different data used in training, showing that our base ST models are representative of current SoTA techniques.

\section{Experimental Settings}

\subsection{Models}
\label{sec:modeling}
\begin{figure*}[t!]
    \centering
    \includegraphics[trim=0 0 0 5,clip,width=1.0\textwidth]{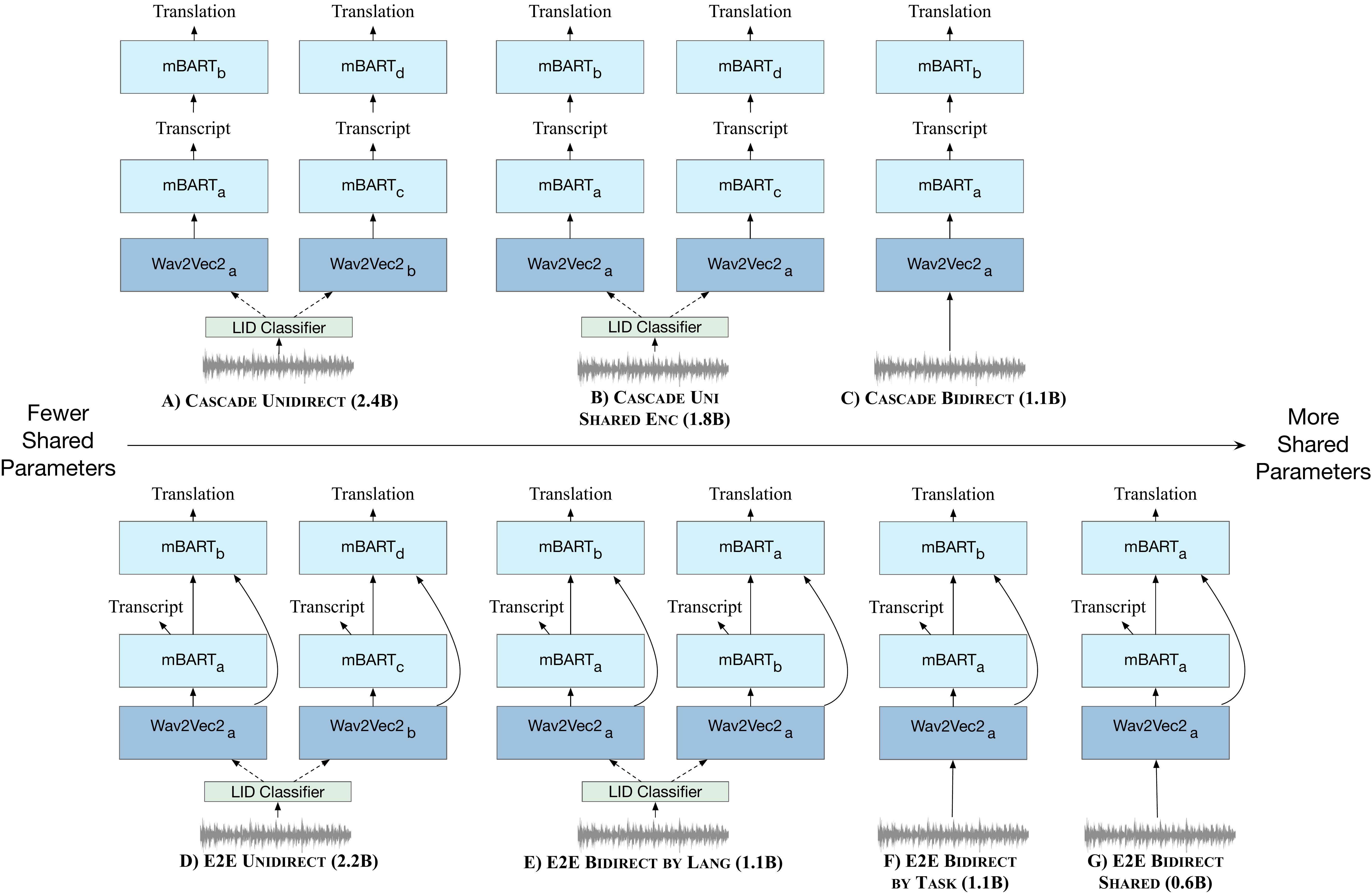}
    \caption{Illustration of model architectures, with cascaded architectures on the top and E2E architectures on the bottom.  Left to right shows the progression of models with the least and the most amount of shared parameters respectively. Subscripts are present to indicate shared modules within each model. Dotted lines indicate a decision where only one path is chosen using the LID. Note that there is no cascade equivalent to the \textsc{Bidirectional E2E Shared} model, as the cascaded model by definition generates transcript then translation separately. The numbers in parentheses stands for the number of model parameters in billions.}
    \label{fig:arch}
\end{figure*}

\paragraph{Joint Transcript/Translation Models}
Many different types of E2E models exist for joint transcript/translation ST \cite{Sperber2020}. Here, we focus on the \textit{triangle} E2E architecture due to its strong performance in previous work \cite{anastasopoulos2018tied,sperber2020consistent}. Following recent work \cite{gallego2021upc,li2020multilingual} we use pre-trained modules as a starting place for our ST model, using a Wav2Vec 2.0 \cite{baevski2020wav2vec} encoder and a mBART 50-50 \cite{liu2020multilingual,tang2020multilingual} decoder. 

Because our task involves joint ASR and ST, we need to adapt the pre-trained decoder to work with the E2E triangle architecture. Specifically, the triangle model's second decoder computes cross attention separately over both the first decoder and the encoder states. We place an additional cross-attention layer after each encoder-attention layer in mBARTs decoder blocks, initializing them with the pre-trained encoder-attention weights. To make sure these weights converge properly, we freeze the entire model for approximately the first epoch while training only the bridge and additional cross attention layers (c.f. Appendix~\ref{app:training}).

As described in Section~\ref{sec:taskanddata}, our task involves modeling intra-sentence CS.  This means that any model used for this task must either explicitly or implicitly learn to model the language of each word in the sentence. Furthermore, as more than one language is being modeled, each sub-component of the model can either be unidirectional or bidirectional. We can thus categorize potential models by how much information is shared within the parameters: the least shared models would be unidirectional and joined together by explicit LID, whereas the most shared would be bidirectional models that learn the LID implicitly. Models and their categorization along this scale are shown in Figure~\ref{fig:arch}.

For cascade models, the most basic would be separate unidirectional cascaded models joined by an LID model. The LID model will explicitly decide what the matrix language is and send the utterance to the model that is best equipped to handle that language (Figure~\ref{fig:arch}A). Note that this approach may suffer from error propagation issues due to incorrect LID. A more parameter-shared version of this model is to make the cascaded model encoder shared between both unidirectional models (Figure~\ref{fig:arch}B).  Finally, we can examine a bidirectional cascade model that shares each component across both languages.  This architecture implicitly learns to model the language of the input, removing the need for an explicit LID model (Figure~\ref{fig:arch}C).

We also examine similar analogues for the E2E triangle model: unidirectional models joined by LID (Figure~\ref{fig:arch}D) and a bidirectional model with LID and a shared encoder (Figure~\ref{fig:arch}E). We can also use the standard triangle model (see \citet{anastasopoulos2018tied} for implementation details) that includes one encoder and two decoders (one for each sub-task) (Figure~\ref{fig:arch}F). Furthermore, we propose to alter the standard triangle model and share both decoder parameters for both languages with a joint bidirectional decoder (Figure~\ref{fig:arch}G, note that the cascade model cannot do this due to the definition of the cascade). By doing so, we hope to provide an inductive bias for the model to more easily handle code-switched data, as the weights of that decoder will already be used to handling multiple languages for both tasks (compared to the bidirectional cascade model, which only shares multilingual parameters for each task of transcript and translation).

\begin{table*}[t!]
\small
    \centering
    \begin{tabular}{l @{\hspace{-0.25\tabcolsep}} rrrr|rrrr}
      \toprule

      \bfseries  &  \multicolumn{4}{c}{\bfseries Not Fine-Tuned} &  \multicolumn{4}{c}{\bfseries Fine-Tuned} \\
      \bfseries  &  \multicolumn{2}{c}{\bfseries  CS} &  \multicolumn{2}{c}{\bfseries  Mono.} &  \multicolumn{2}{c}{\bfseries  CS} &  \multicolumn{2}{c}{\bfseries  Mono.} \\
  
      \bfseries Models &  \multicolumn{1}{c @{\hspace{0.125\tabcolsep}}}{\bfseries $\downarrow$ WER} & \multicolumn{1}{c @{\hspace{0.125\tabcolsep}}}{\bfseries  $\uparrow$ BLEU}  &  \multicolumn{1}{c @{\hspace{0.125\tabcolsep}}}{\bfseries $\downarrow$ WER} & \multicolumn{1}{c @{\hspace{0.125\tabcolsep}}}{\bfseries  $\uparrow$ BLEU}  &  \multicolumn{1}{c @{\hspace{0.125\tabcolsep}}}{\bfseries $\downarrow$ WER} & \multicolumn{1}{c @{\hspace{0.125\tabcolsep}}}{\bfseries  $\uparrow$ BLEU}  &  \multicolumn{1}{c @{\hspace{0.125\tabcolsep}}}{\bfseries $\downarrow$ WER} & \multicolumn{1}{c @{\hspace{0.125\tabcolsep}}}{\bfseries  $\uparrow$ BLEU}\\
      \midrule

\textsc{Cascade Unidirect}      &    37.1 &    22.5 &    26.6 &    24.7 &    33.5 &    24.6 &    24.8 &    25.5 \\
                    &   (-0.8) & (-0.4) & (-3.1) &	(+0.9) &	(-0.4) &	(0.0)	& (-1.0) &	(+0.2) \\
\textsc{Cascade Uni Shared Enc} &    36.0 &    21.6 &    25.6 &    24.3 &    31.2 &    25.4 &    25.6 &    24.8 \\
                   &  (0.0) &	(+0.6) &	(0.0) &	(+0.5) &	(+0.1) &	(+0.2)	 & (-0.3) &	(+0.1) \\
\textsc{E2E Unidirect}          &    36.6 &    22.3 &    26.7 &    25.0 &    33.4 &    24.4 &    25.3 &    25.5 \\
                       &  (-0.9) &	(-0.1) &	(-3.5) &	(+1.0)	 & (-0.2) &	(+0.1) &	(-1.4) &	(+0.4) \\
\textsc{E2E Bidirect by Lang}   &    37.0 &    23.4 &    27.2 &    25.0 &    36.7 &    22.8 &    27.3 &    25.0 \\
                       &  (-0.9) &	(-0.1) &	(-1.9) &	(+0.5) &	(-0.8) &	(+0.2) &	(-2.0) &	(+0.4) \\
      \bottomrule
    \end{tabular}
    \caption{Comparison of Oracle vs Predicted LID results on the Fisher dataset. Numbers in parenthesis are the difference to the corresponding model with oracle LID. Note that the \textbf{Oracle LID improves upon the Predicted LID} in most cases. Conclusions are similar for the Miami corpus (see Appendix~\ref{app:all_lid} Table~\ref{tab:all_lid})}
    \label{tab:fisher_lid}
\end{table*}

\paragraph{Language Identification Model}
\label{sec:lid}
We train the language identification (LID) model to identify the matrix language. For consistency with our other models (and similar to concurrent work, e.g. \citet{tjandra2021improved}), we use a pre-trained Wav2Vec2 along with a classifier layer to predict whether the utterance is majority Spanish or majority English. We train the model in the same fashion as the joint transcription and translation models (Section~\ref{sec:modeling} and Appendix~\ref{app:training}) but train on the LID data instead.

The data for the LID model was gathered by taking the CS data\footnote{For the No-FT case (Section~\ref{sec:training_process}), we exclude the CS data when training the LID model.} from the training set of the Fisher corpus and combining it with randomly sampled data from several different datasets in order to help the model learn despite the domain of the audio. We use MuST-C English audio, CoVoST English audio, CoVoST Spanish audio, and the monolingual Spanish audio from the training sets of Fisher and Miami. We found that upsampling the CS training set by 2 and using the same amount of data (2x the number of the CS set) for CoVoST and MuST-C provided the best results: 98\%+ accuracy on CoVoST and MuST-C, 89\% on the Fisher CS validation and test sets, and 72\% on the Miami CS test set (due to the noisy data). As a large proportion of the CS data is close to 50\% code-switched (see Figure~\ref{fig:cs_distributions}), it becomes more difficult for the model to predict the matrix language correctly.

\begin{table*}[t!]
\small
    \centering
    \begin{tabular}{ll @{\hspace{-0.25\tabcolsep}} rrrr|rrrr}
      \toprule
      \bfseries & & \multicolumn{4}{c}{\bfseries Not Fine-Tuned} &  \multicolumn{4}{c}{\bfseries Fine-Tuned} \\
        \bfseries & & \multicolumn{2}{c}{\bfseries CS} &  \multicolumn{2}{c}{\bfseries Mono.} &  \multicolumn{2}{c}{\bfseries CS} &  \multicolumn{2}{c}{\bfseries Mono.} \\
      \bfseries & Model &  \multicolumn{1}{c @{\hspace{0.125\tabcolsep}}}{\bfseries $\downarrow$ WER} & \multicolumn{1}{c| @{\hspace{0.125\tabcolsep}}}{\bfseries $\uparrow$ BLEU} &  \multicolumn{1}{c @{\hspace{0.125\tabcolsep}}}{\bfseries $\downarrow$ WER} & \multicolumn{1}{c| @{\hspace{0.125\tabcolsep}}}{\bfseries $\uparrow$ BLEU} &  \multicolumn{1}{c @{\hspace{0.125\tabcolsep}}}{\bfseries $\downarrow$ WER} & \multicolumn{1}{c| @{\hspace{0.125\tabcolsep}}}{\bfseries $\uparrow$ BLEU} &  \multicolumn{1}{c @{\hspace{0.125\tabcolsep}}}{\bfseries $\downarrow$ WER} & \multicolumn{1}{c @{\hspace{0.125\tabcolsep}}}{\bfseries $\uparrow$ BLEU} \\
      \midrule
 \parbox[t]{2mm}{\multirow{7}{*}{\rotatebox[origin=c]{90}{Fisher}}} &
\textsc{Cascade Unidirect}      &  37.1 &  22.5 & 26.6 &  24.7 &  33.5 &  24.6 &  24.8 &  25.5 \\
& \textsc{Cascade Uni Shared Enc} &  36.0 &  21.6 &   25.6 &   24.3  &  31.2 &  *25.4 &  25.6 &  24.8 \\
& \textsc{Cascade Bidirect}       &  37.2 &  21.8 &  26.5 &  24.1 &  33.2 &  23.2 &  28.1 &  23.2 \\
& \textsc{E2E Unidirect}          & 36.6 &  22.3 &  26.7 & 25.0 &  33.4 &  24.4 &  25.3 &  25.5 \\
& \textsc{E2E Bidirect by Lang}   & 37.0 &  \textbf{23.4} & 27.2 &  25.0 &  36.7 &  22.8 &  27.3 &  25.0 \\
& \textsc{E2E Bidirect by Task}   &  *34.1 &  *23.0 &  23.6 &  26.0 &  *30.1 &  \textbf{25.6} &  *24.3 &  25.6 \\
& \textsc{E2E Bidirect Shared}    &  \textbf{33.8} &  *23.3 &  \textbf{23.2} &  \textbf{26.2} &  \textbf{30.0} &  *25.4 &  \textbf{24.1} &  \textbf{26.1} \\
\midrule
 \parbox[t]{2mm}{\multirow{7}{*}{\rotatebox[origin=c]{90}{Miami}}} &
\textsc{Cascade Unidirect}      &  65.2 &   8.8 &   52.3 &   16.8  &  64.8 &  10.8 &  \textbf{51.5} &  16.8 \\
& \textsc{Cascade Uni Shared Enc} &  60.2 &   9.7 &   53.8 &   15.7 &  55.0 &  \textbf{14.7} &  55.6 &  15.3 \\
& \textsc{Cascade Bidirect}       &  61.4 &   9.3 &  54.0 &  14.8 &  57.4 &  10.6 &  58.2 &  14.0 \\
& \textsc{E2E Unidirect}          & 65.6 &   10.1 &  53.0 &  17.2 &  65.1 &  11.7 &  *51.4 &  \textbf{17.6} \\
& \textsc{E2E Bidirect by Lang}   &   69.5 &   \textbf{12.4} &    55.2 &   16.5  &  69.3 &  11.5 &  54.5 &  16.6 \\
& \textsc{E2E Bidirect by Task}   &  59.9 &  11.0 &  *50.0 &  *18.1 &  *53.6 &  *13.8 &  52.6 &  *17.5 \\
& \textsc{E2E Bidirect Shared}    &  \textbf{58.9} &  *11.8 &  \textbf{49.9} &  \textbf{18.3} &  \textbf{53.0} &  *14.1 &  52.1 &  *17.4 \\
      \bottomrule
    \end{tabular}
       \caption{Test set scores, with results from the Fisher corpus on the top half and the Miami corpus on the bottom half. Bold scores indicate the best score in the column, while asterisks indicate results that are statistically similar to the best score in the column group using a bootstrap resampling test with $\alpha=0.05$. }
    \label{tab:main_results}
\end{table*}

\subsection{Training Process and Evaluation}
\label{sec:training_process}
For all dataset evaluations, we use word error rate (WER) and character error rate (CER) for the transcript and Charcut (CCT) \cite{lardilleux2017charcut} and sacreBLEU \cite{post2018call} for the translation. However, we found that there was no difference in conclusions between each of the two metrics (WER vs CER and BLEU vs Charcut) and thus we only report BLEU/WER in the main text (see Appendix~\ref{app:training} for implementation details). For tables showing all metrics, see Appendix~\ref{app:all}.

We evaluate our models on the Fisher and Miami test sets (with both CS-only and monolingual-only test sets) in two different settings: (1) without fine-tuning them on CS data (No-FT) and (2) after fine-tuning the already trained ST models on the Fisher CS Training set (FT). For models consisting of two monolingual sub-models we fine-tune both on the CS data. During fine-tuning we employ the same hyperparameters as in the original experiment, but perform early stopping on the Fisher CS Dev set. We use significance tests to verify the reliability of our results \cite{Koehn2004}. We run bootstrap resampling tests against the best performing model, using $\alpha=0.05$. More training parameters such as learning rates, etc. can be found in Appendix \ref{app:training}.

\section{Results}

\subsection{Scores on Test Sets}
In this section, we explore the results of doing ST for CS data along the two axes of unidirectional vs bidirectional and end-to-end vs cascade.

We see results for models using explicit LID prediction in Table~\ref{tab:fisher_lid}, showing that models that use the \textbf{predicted LID perform worse than those that use Oracle LID} (e.g. 36.6 vs 35.7 WER for the \textsc{E2E Unidirect}). This provides a slight advantage for the bidirectional models that learn LID implicitly. However, the predicted LID case is the realistic setting, and thus we use it for the remainder of our experiments.

\begin{figure}[t]
    \centering
    \includegraphics[trim=15 15 0 10,width=0.45\textwidth]{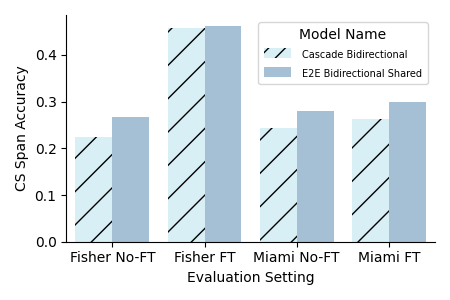}
    \caption{Accuracy of the models in generating the CS spans. Note that this excludes all non-exact matches and is a lower bound on performance.}
    \label{fig:cs_word_acc}
\end{figure}

\begin{table*}[t!]
    \centering
    \setlength\extrarowheight{1pt}
    \small
    \begin{tabular}{lp{6.5cm}p{6.5cm}}
      Model & Transcript & Translation \\
    \Xhline{2\arrayrulewidth}
     Reference & \textcolor{paperblue}{si entonces volví aquí a la casa si el} \textcolor{paperred}{fall break} & \textcolor{paperblue}{yes so I returned here to the house yes the} \textcolor{paperred}{fall break} \\
   \hline
   Cascade & \textcolor{paperblue}{si entonces volví aquí a la casa si es} \textcolor{paperred}{folvereak} & \textcolor{paperblue}{yes then I returned here at home yes its} \textcolor{paperred}{folvereak} \\
  \hline
  E2E & \textcolor{paperblue}{si entonces volví aquí a la casa si es} \textcolor{paperred}{fallbreak} & \textcolor{paperblue}{yes so I came back to the house yes its} \textcolor{paperred}{fallbreak} \\
      \Xhline{2\arrayrulewidth}
    \end{tabular}
    \caption{Example generated output from the \textsc{Cascade Bidirect} and \textsc{E2E Bidirect Shared} models. Note the error propagation in the cascade model.}
    \label{fig:errors}
\end{table*}

When we examine the models along the scale of unidirectional to bidirectional, we see that \textbf{higher amounts of shared parameters are correlated with higher scores}, e.g. bidirectional is better. We see that on all datasets and evaluation settings (Table~\ref{tab:main_results}) that the \textsc{E2E Bidirect Shared} model is either statistically similar or outperforms all other models, except for the Miami Monolingual FT case, where it comes in 3rd. Thus, the inductive bias of sharing the multilingual task parameters provides a gain of approximately 3.5 WER points (33.8 vs 37.3) and 1.5 BLEU points (23.3 vs 21.9) for the \textsc{E2E Bidirect Shared} model over the \textsc{E2E Unidirect} model on the Fisher dataset, with similar performance on the Miami dataset.

We can also examine Table~\ref{tab:main_results} to see how the cascade models compare to the E2E models. The results show that the \textbf{cascaded models perform the same or worse than the E2E models} they compare to w.r.t. parameter sharing, with the best overall model being the \textsc{E2E Bidirect Shared}, beating the \textsc{Cascade Bidirect} (e.g. 33.8 vs 37.2 WER or 23.3 vs 21.8 BLEU on Fisher No-FT).

Table~\ref{tab:main_results} also illustrates that  \textbf{fine-tuning models on CS data improves scores} on CS test sets (33.8 vs 30.0 WER for the \textsc{E2E Bidirect Shared} on Fisher, 58.9 vs 53.0 for Miami). These gains are consistent for the Fisher dataset, which is the domain of the CS training set, however there are still gains for the out-of-domain Miami CS data. These results suggest that additional pre-training on natural or synthetic data (in both audio/text modalities) would likely be fruitful future work. When we examine how fine-tuning on CS data changes the model's monolingual scores, we find that they generally improve the monolingual results for the unidirectional models, but tend to make bidirectional models slightly worse, perhaps due to interference between the languages and tasks in the same weights. However, overall we find that fine-tuning provides large gains for CS with only minor decreases in monolingual performance.

\subsection{Model Analysis}
\label{sec:results_analysis}
We also provide further analysis of the CS output of the best model and its cascaded counterpart (\textsc{Bidirect Cascade} and  \textsc{E2E Bidirect Shared}). We perform three analyses: (1) comparing utterance level scores vs the proportion of CS words in the utterance, (2) computing the exact match accuracy of the CS spans in the model's output, and (3) qualitatively examining model output. 

We check the correlation between the proportion of CS words in a sentence and the model's score, using a linear model to find the $R^2$ values. We found that surprisingly, there was \textbf{no correlation between the proportion of CS words and the models score} for any of the different models or metrics ($R^2 < 0.025$ for all models and metrics). A graphical depiction of the model's scores over CS proportions is in the Appendix, Figure~\ref{fig:charcut_vs_prop}. We note that this finding was the same for comparing the \textit{number} of CS words instead of the \textit{proportion}. This finding implies that the models are surprisingly robust to the amount of CS in a sentence.

Although BLEU and WER scores show how well the models do on the CS data, we can further isolate the performance of these models on only the code-switched parts of the utterances. To do so, we isolate all CS spans in the sentences and check to see if the model's output contains the exact-match of those spans. We note that this metric does not take into account synonyms or different tenses of the same word, making it a stricter metric serving as a lower bound of absolute performance. We see in Figure~\ref{fig:cs_word_acc} that the E2E model still outperforms the cascade on CS spans, with Fisher No-FT scores around 20-30\% and Fisher FT scores around 45\%.

Finally, we can also examine the model's outputs. We inspected 200 output sentences for the monolingual subsets and found that both models generated the correct language in every case, indicating that they correctly learned the implicit LID. However, we can see that the cascade model does struggle with error propagation (especially so in the CS setting, Table~\ref{fig:errors}), likely causing part of the difference between the E2E and cascade models. 

Although the CS WER and BLEU scores are not as high as they are on cleaner monolingual datasets such as CoVoST (Appendix~\ref{app:training}), their performance is competitive with their respective monolingual performance on Miami and Fisher, even in the No-FT setting. We believe that with additional data and improvements ST models will be well-equipped to handle CS in practical situations and that \textbf{overall, models show strong CS performance}.

\section{Conclusion}
In this work, we expand the ST literature to explore code-switching, contributing a new task framework for ST that extends the joint transcription and translation setup.
To further progress, we built and open-sourced a new ST corpus for CS from existing public datasets.
We evaluated a range of models, showing that using bilingual joint decoders provides gains over using separate task decoders.
We also showed that E2E systems provide better performance than their cascading counterparts on the CS task. 
Overall, our work shows that ST models can perform well on CS applications with both no fine-tuning and in low-resource settings, opening the door to new and exciting areas of future work.
\bibliography{anthology,acl}
\bibliographystyle{acl_natbib}

\clearpage
\appendix

\begin{table}
    \centering
    \small
    \begin{tabular}{l @{\hspace{-0.25\tabcolsep}} rr|rr}
      \toprule
      \bfseries  &  \multicolumn{2}{c}{\bfseries CS} &  \multicolumn{2}{c}{\bfseries Mono.} \\
      \bfseries Models &  \multicolumn{1}{c @{\hspace{0.125\tabcolsep}}}{\bfseries $\downarrow$ WER}  & \multicolumn{1}{c| @{\hspace{0.125\tabcolsep}}}{\bfseries $\uparrow$ BLEU} & \multicolumn{1}{c @{\hspace{0.125\tabcolsep}}}{\bfseries $\downarrow$ WER} & \multicolumn{1}{c @{\hspace{0.125\tabcolsep}}}{\bfseries $\uparrow$ BLEU}\\
      \midrule
 Random Init &  69.6 &  11.0 &  59.6 &  13.2 \\
 Pre-trained &  33.8 &  23.3 &  23.2 &  26.2 \\
      \bottomrule
    \end{tabular}
    \caption{Comparison of the E2E bidirectional shared model with pre-training vs random initialization on the Fisher code-switched test sets.}
    \label{tab:pretrained}
\end{table}

\section{Training and Evaluation Details}
\label{app:training}
We follow \citet{gallego2021upc,li2020multilingual} and use a triangular learning rate, adapting the step count to depend on the batch size (as not all models could fit the same batch size) with (64 / batch size) * 500 warm up steps, (64 / batch size) * 500 hold steps, (64 / batch size) * 3000 decay steps, a beta of 0.9, and a beta2 of 0.98. The learning rate was selected from running a search over \{0.01, 0.005, 0.001, 0.0005, 0.0001, 0.0005\}. We found that 0.0005 was best for all models, so we examined learning rates again between 0.0001 to 0.001 (by 0.0001) and found that they all performed similarly, thus we use 0.0005 in our experiments. For efficiency in batch size while training, we removed all instances whose audio length was longer than 20 seconds. We freeze the attention layers for the first 500 * (64 / batch size) steps, which is approximately the first epoch of training.

We initially trained the models on only CoVoST and MuST-C and found there was a large domain shift between these datasets and the comparatively noisier Fisher and Miami datasets. As domain shift was not the focus of this paper, we further trained the models on the Fisher and Miami monolingual training sets to reduce the effect of domain shift. 

As a sanity check of the effectiveness of our training, we also include scores in Table \ref{tab:training} for the test sets of CoVoST and MuST-C. We note that our scores are close to the SoTA scores of \citet{li2020multilingual} on CoVoST (and they use the large Wav2Vec2 model while we use the base version) and our MuST-C scores are higher than that of \citet{gallego2021upc}.

We evaluate using word error rate, character error rate, charcut, and BLEU. As the models learn different punctuation techniques from a variety of sources, including MuST-C, CoVoST, Miami, and Fisher, we remove all punctuation from the output before evaluating on the CS/Mono test sets, in order to only measure scores on the content. For BLEU, we use SacreBLEU with parameters case.lc+numrefs.1+smooth.4.0+tok.13a. 

\begin{figure}[t]
    \centering
    \includegraphics[trim=15 15 0 10,width=0.47\textwidth]{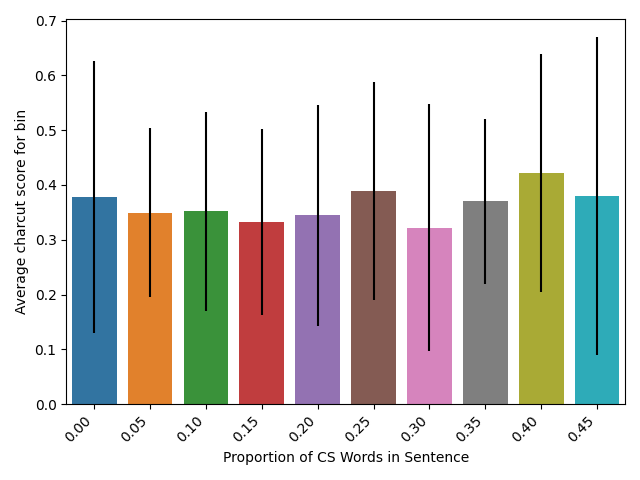}
    \caption{Charcut performance of the \textsc{E2E Bidirect Shared} model on sentences with various levels of CS proportions. Note that there is no clear correlation, as described in Section~\ref{sec:results_analysis}. Black lines indicate error bars of 2 standard deviations while the bar represents the average.}
    \label{fig:charcut_vs_prop}
\end{figure}

\section{More LID Comparisons}
\label{app:all_lid}
We show results for all models that use LID on both datasets in Table~\ref{tab:all_lid}. Note the conclusions remain the same as Table~\ref{tab:fisher_lid}.

\begin{table*}[t!]
    \small
    \centering
    \begin{tabular}{ll @{\hspace{-0.25\tabcolsep}} rrrr|rrrr}
      \toprule
      \bfseries & & \multicolumn{4}{c}{\bfseries No Fine-Tuning} &  \multicolumn{4}{c}{\bfseries Fine-Tuned} \\
        \bfseries & & \multicolumn{2}{c}{\bfseries CS} &  \multicolumn{2}{c}{\bfseries Mono.} &  \multicolumn{2}{c}{\bfseries CS} &  \multicolumn{2}{c}{\bfseries Mono.} \\
      \bfseries & Model &  \multicolumn{1}{c @{\hspace{0.125\tabcolsep}}}{\bfseries $\downarrow$ WER} & \multicolumn{1}{c| @{\hspace{0.125\tabcolsep}}}{\bfseries $\uparrow$ BLEU} &  \multicolumn{1}{c @{\hspace{0.125\tabcolsep}}}{\bfseries $\downarrow$ WER} & \multicolumn{1}{c| @{\hspace{0.125\tabcolsep}}}{\bfseries $\uparrow$ BLEU} &  \multicolumn{1}{c @{\hspace{0.125\tabcolsep}}}{\bfseries $\downarrow$ WER} & \multicolumn{1}{c| @{\hspace{0.125\tabcolsep}}}{\bfseries $\uparrow$ BLEU} &  \multicolumn{1}{c @{\hspace{0.125\tabcolsep}}}{\bfseries $\downarrow$ WER} & \multicolumn{1}{c @{\hspace{0.125\tabcolsep}}}{\bfseries $\uparrow$ BLEU} \\
      \midrule
   \parbox[t]{2mm}{\multirow{8}{*}{\rotatebox[origin=c]{90}{Fisher}}} &
   
  \textsc{Cascade Unidirect}     &    37.1 &    22.5 &    26.6 &    24.7 &    33.5 &    24.6 &    24.8 &    25.5 \\
 &                       &  (36.3) &  (22.1) &  (23.5) &  (25.6) &  (33.1) &  (24.6) &  (23.8) &  (25.7) \\
& \textsc{E2E Unidirect}          &    36.6 &    22.3 &    26.7 &    25.0 &    33.4 &    24.4 &    25.3 &    25.5 \\
 &                       &  (35.7) &  (22.2) &  (23.2) &  (26.0) &  (33.2) &  (24.5) &  (23.9) &  (25.9) \\
& \textsc{Cascade Uni Shared Enc}  &    36.0 &    21.6 &    25.6 &    24.3 &    31.2 &    25.4 &    25.6 &    24.8 \\
&                        &  (36.0) &  (22.2) &  (25.6) &  (24.8) &  (31.3) &  (25.6) &  (25.3) &  (24.9) \\
& \textsc{E2E Bidirect by Lang}   &    37.0 &    23.4 &    27.2 &    25.0 &    36.7 &    22.8 &    27.3 &    25.0 \\
 &                       &  (36.1) &  (23.3) &  (25.3) &  (25.5) &  (35.9) &  (23.0) &  (25.3) &  (25.4) \\

\midrule
\parbox[t]{2mm}{\multirow{8}{*}{\rotatebox[origin=c]{90}{Miami}}} &
 \textsc{Cascade Unidirect}       &    65.2 &     8.8 &    52.3 &    16.8 &    64.8 &    10.8 &    51.5 &    16.8 \\
 &                       &  (61.4) &   (8.3) &  (50.0) &  (17.3) &  (64.4) &  (10.8) &  (50.9) &  (16.9) \\
& \textsc{E2E Unidirect}          &    65.6 &    10.1 &    53.0 &    17.2 &    65.1 &    11.7 &    51.4 &    17.6 \\
  &                      &  (63.1) &   (9.4) &  (51.2) &  (17.7) &  (65.6) &  (11.7) &  (50.7) &  (17.7) \\
& \textsc{Cascade Uni Shared Enc} &    60.2 &     9.7 &    53.8 &    15.7 &    55.0 &    14.7 &    55.6 &    15.3 \\
&                        &  (60.2) &   (8.8) &  (53.8) &  (16.0) &  (56.0) &  (14.4) &  (55.5) &  (15.3) \\
& \textsc{E2E Bidirect by Lang}  &    69.5 &    12.4 &    55.2 &    16.5 &    69.3 &    11.5 &    54.5 &    16.6 \\
 &                       &  (69.7) &  (10.7) &  (53.4) &  (16.7) &  (69.7) &  (10.4) &  (53.2) &  (16.6) \\
      \bottomrule
    \end{tabular}
    \caption{Scores on the code-switched test sets for the models using LID, with results from zero CS training on the left and results after fine-tuning on the right.\label{tab:all_lid}}
\end{table*}

\section{Random Initialization Results}
\label{app:pretraining}
We also perform an ablation of these pre-trained scores (Table~\ref{tab:pretrained}) for the \textsc{E2E Bidirect Shared} model, as it is the best performing model overall. We tried many different setups for training it from from scratch rather than loading the pre-trained weights. We found that it was very difficult for this model to converge, and when it did, the results were sub-par.

\section{Training Results}
\label{app:training_test}
We include the scores of evaluating our models on the test sets of the ST training data (MuST-C and CoVoST) in Table~\ref{tab:training}. We also include the results of fine-tuning performance on the CS dev set in Table~\ref{tab:cs_dev_training}, which roughly mirrors the main results. 

\section{Expanded Results}
\label{app:all}
For brevity, we do not include the CER and Charcut metrics in the main text.  In this section we included tables with all metrics for all results (Table~\ref{tab:miami_all} for Miami and Table~\ref{tab:fisher_all} for Fisher). We note however, that the WER and BLEU scores align with the CER and Charcut scores, and thus our conclusions remain the same.

\begin{table*}[t!]
    \centering
    \begin{tabular}{l @{\hspace{-0.25\tabcolsep}} rrrr}
      \toprule
      \bfseries  &  \multicolumn{4}{c}{\bfseries Fisher CS Dev Set} \\
      \bfseries Models &  \multicolumn{1}{c @{\hspace{0.125\tabcolsep}}}{\bfseries $\downarrow$ WER} & \multicolumn{1}{c @{\hspace{0.125\tabcolsep}}}{\bfseries $\downarrow$ CER} & \multicolumn{1}{c @{\hspace{0.125\tabcolsep}}}{\bfseries $\downarrow$ CCT} & \multicolumn{1}{c @{\hspace{0.125\tabcolsep}}}{\bfseries $\uparrow$ BLEU} \\
\midrule
          \textsc{Cascade Unidirect} & 34.2 & 19.3 & 38.4 &  26.4 \\
              \textsc{E2E Unidirect} & 33.0 & 18.9 & 37.3 &  27.8 \\
     \textsc{Cascade Uni Shared Enc} & 32.3 & 17.9 & 38.4 &  24.9 \\
       \textsc{E2E Bidirect by Lang} & 36.3 & 23.0 & 39.3 &  26.3 \\
       \textsc{E2E Bidirect by Task} & 31.1 & 17.0 & 35.1 &  29.0 \\
           \textsc{Cascade Bidirect} & 35.1 &  19.2 & 39.7 & 23.8 \\
        \textsc{E2E Bidirect Shared} & 31.7 & 17.5 & 35.2 &  28.3 \\

\bottomrule
\end{tabular}
\caption{Scores on the Fisher CS Dev set. CCT stands for Charcut. Note that this mirrors the main results in Table~\ref{tab:main_results}.\label{tab:cs_dev_training}}
\end{table*}

\begin{table*}[t!]
\small
    \centering
    \begin{tabular}{l @{\hspace{-0.25\tabcolsep}} rrrr|rrrr}
      \toprule
      \bfseries  &  \multicolumn{4}{c}{\bfseries MuST-C Test Set} &  \multicolumn{4}{c}{\bfseries CoVoST Test Set} \\
      \bfseries Models &  \multicolumn{1}{c @{\hspace{0.125\tabcolsep}}}{\bfseries $\downarrow$ WER} & \multicolumn{1}{c @{\hspace{0.125\tabcolsep}}}{\bfseries $\downarrow$ CER} & \multicolumn{1}{c @{\hspace{0.125\tabcolsep}}}{\bfseries $\downarrow$ CCT} & \multicolumn{1}{c| @{\hspace{0.125\tabcolsep}}}{\bfseries $\uparrow$ BLEU} & \multicolumn{1}{c @{\hspace{0.125\tabcolsep}}}{\bfseries $\downarrow$ WER} & \multicolumn{1}{c @{\hspace{0.125\tabcolsep}}}{\bfseries $\downarrow$ CER} & \multicolumn{1}{c @{\hspace{0.125\tabcolsep}}}{\bfseries $\downarrow$ CCT} & \multicolumn{1}{c @{\hspace{0.125\tabcolsep}}}{\bfseries $\uparrow$ BLEU} \\
\toprule
          \textsc{Cascade Unidirect} & 11.2 &  7.6 & 36.3 &  29.4 & 17.2 &  5.8 & 35.6 &  26.9 \\
              \textsc{E2E Unidirect} & 13.0 &  8.9 & 37.3 &  27.8 & 18.6 &  6.4 & 36.0 &  26.2 \\
     \textsc{Cascade Uni Shared Enc} & 12.0 &  8.1 & 37.7 &  26.9 &  22.9 &  7.3 & 36.2 &  26.0 \\
       \textsc{E2E Bidirect by Lang} & 11.6 &  7.8 & 36.6 &  28.6 & 19.7 &  7.7 & 37.1 &  25.4 \\
       \textsc{E2E Bidirect by Task} & 11.4 &  7.6 & 36.6 &  28.4 & 17.9 &  6.0 & 35.3 &  26.8 \\
           \textsc{Cascade Bidirect} &  13.6 &  9.5 &  39.7 &   24.5 &  22.9 &  7.3 &  38.8 &   22.8 \\
        \textsc{E2E Bidirect Shared} & 11.6 &  7.7 & 36.6 &  28.5 & 18.1 &  6.2 & 35.0 &  27.4 \\
\bottomrule
\end{tabular}
\caption{Scores on the MustC and CovoST datasets. CCT stands for Charcut.\label{tab:training}}
\end{table*}

\begin{table*}[t!]
    \centering
    \small
    \begin{tabular}{ll @{\hspace{-0.25\tabcolsep}} rrrr|rrrr}
      \toprule
      & \bfseries  &  \multicolumn{8}{c}{\bfseries Miami} \\
      & \bfseries  &  \multicolumn{4}{c}{\bfseries CS Test Set} &  \multicolumn{4}{c}{\bfseries Monolingual Test Set} \\
      & \bfseries Models &  \multicolumn{1}{c @{\hspace{0.125\tabcolsep}}}{\bfseries $\downarrow$ WER} & \multicolumn{1}{c @{\hspace{0.125\tabcolsep}}}{\bfseries $\downarrow$ CER} & \multicolumn{1}{c @{\hspace{0.125\tabcolsep}}}{\bfseries $\downarrow$ CCT} & \multicolumn{1}{c| @{\hspace{0.125\tabcolsep}}}{\bfseries $\uparrow$ BLEU} & \multicolumn{1}{c @{\hspace{0.125\tabcolsep}}}{\bfseries $\downarrow$ WER} & \multicolumn{1}{c @{\hspace{0.125\tabcolsep}}}{\bfseries $\downarrow$ CER} & \multicolumn{1}{c @{\hspace{0.125\tabcolsep}}}{\bfseries $\downarrow$ CCT} & \multicolumn{1}{c @{\hspace{0.125\tabcolsep}}}{\bfseries $\uparrow$ BLEU} \\
\toprule
         \parbox[t]{2mm}{\multirow{11}{*}{\rotatebox[origin=c]{90}{Not Fine-Tuned}}} &

                \textsc{Cascade Unidirect} & 63.6 & 43.3 & 65.2 &   8.7 & 52.3 & 34.1 & 51.9 &  17.0 \\
&          \textsc{Cascade Unidirect Ora.} & 61.4 & 41.6 & 67.4 &   8.3 & 50.0 & 32.4 & 50.9 &  17.3 \\
 &                    \textsc{E2E Unidirect} & 64.0 & 43.0 & 64.0 &   9.9 & 53.0 & 34.6 & 51.0 &  17.4 \\
&              \textsc{E2E Unidirect Ora.} & 63.1 & 42.2 & 66.5 &   9.4 & 51.2 & 33.1 & 50.1 &  17.7 \\
&           \textsc{Cascade Uni Shared Enc} & 60.2 & 39.7 & 63.7 &   9.3 & 53.8 & 34.1 & 52.7 &  15.9 \\
&   \textsc{Cascade Uni Shared Enc Ora.} & 60.2 & 39.7 & 66.1 &   8.8 & 53.8 & 34.1 & 52.2 &  16.0 \\
&              \textsc{E2E Bidirect by Lang} & 68.8 & 48.4 & 61.2 &  11.5 & 54.6 & 37.3 & 52.0 &  16.7 \\
&       \textsc{E2E Bidirect by Lang Ora.} & 69.7 & 49.4 & 63.6 &  10.7 & 53.4 & 36.3 & 51.5 &  16.7 \\
&              \textsc{E2E Bidirect by Task} & 59.9 & 39.6 & 59.4 &  11.0 & 50.0 & 32.6 & 49.7 &  18.1 \\
&                  \textsc{Cascade Bidirect} & 61.4 & 39.8 & 62.2 &   9.3 & 54.0 & 34.1 & 53.1 &  14.8 \\
&               \textsc{E2E Bidirect Shared} & 58.9 & 39.1 & 58.5 &  11.8 & 49.9 & 32.2 & 49.3 &  18.3 \\

      \midrule
         \parbox[t]{2mm}{\multirow{11}{*}{\rotatebox[origin=c]{90}{Fine-Tuned}}} &

                \textsc{Cascade Unidirect} & 64.8 & 42.0 & 56.5 &  10.8 & 51.5 & 33.4 & 51.1 &  16.8 \\
&       \textsc{Cascade Unidirect Ora.} & 64.4 & 41.8 & 56.4 &  10.8 & 50.9 & 32.9 & 50.6 &  16.9 \\
&                     \textsc{E2E Unidirect} & 65.1 & 43.0 & 56.9 &  11.7 & 51.4 & 33.7 & 50.4 &  17.6 \\
&              \textsc{E2E Unidirect Ora.} & 65.6 & 43.1 & 57.0 &  11.7 & 50.7 & 33.2 & 49.9 &  17.7 \\
&           \textsc{Cascade Uni Shared Enc} & 55.0 & 35.2 & 51.4 &  14.7 & 55.6 & 35.9 & 52.9 &  15.3 \\
&     \textsc{Cascade Uni Shared Enc Ora.} & 56.0 & 35.7 & 51.7 &  14.4 & 55.5 & 35.9 & 52.7 &  15.3 \\
&              \textsc{E2E Bidirect by Lang} & 69.3 & 48.6 & 61.3 &  11.5 & 54.5 & 37.2 & 52.1 &  16.6 \\
&       \textsc{E2E Bidirect by Lang Ora.} & 69.7 & 49.5 & 63.8 &  10.4 & 53.2 & 36.1 & 51.5 &  16.6 \\
&              \textsc{E2E Bidirect by Task} & 53.6 & 35.0 & 53.3 &  13.8 & 52.6 & 34.4 & 50.5 &  17.5 \\
&                  \textsc{Cascade Bidirect} & 57.4 & 36.3 & 58.8 &  10.6 & 58.2 & 36.6 & 55.1 &  14.0 \\
&               \textsc{E2E Bidirect Shared} & 53.0 & 35.0 & 54.4 &  14.1 & 52.1 & 33.9 & 50.4 &  17.4 \\

\bottomrule
\end{tabular}
\caption{Scores on the Miami dataset. CCT stands for Charcut. Results from zero CS training are on the top half and results after fine-tuning are on the bottom half. Ora stands for Oracle. \label{tab:miami_all}}
\end{table*}

\begin{table*}[t!]
    \centering
    \small
    \begin{tabular}{ll @{\hspace{-0.25\tabcolsep}} rrrr|rrrr}
      \toprule
      & \bfseries  &  \multicolumn{8}{c}{\bfseries Fisher} \\
      & \bfseries  &  \multicolumn{4}{c}{\bfseries CS Test Set} &  \multicolumn{4}{c}{\bfseries Monolingual Test Set} \\
      & \bfseries Models &  \multicolumn{1}{c @{\hspace{0.125\tabcolsep}}}{\bfseries $\downarrow$ WER} & \multicolumn{1}{c @{\hspace{0.125\tabcolsep}}}{\bfseries $\downarrow$ CER} & \multicolumn{1}{c @{\hspace{0.125\tabcolsep}}}{\bfseries $\downarrow$ CCT} & \multicolumn{1}{c| @{\hspace{0.125\tabcolsep}}}{\bfseries $\uparrow$ BLEU} & \multicolumn{1}{c @{\hspace{0.125\tabcolsep}}}{\bfseries $\downarrow$ WER} & \multicolumn{1}{c @{\hspace{0.125\tabcolsep}}}{\bfseries $\downarrow$ CER} & \multicolumn{1}{c @{\hspace{0.125\tabcolsep}}}{\bfseries $\downarrow$ CCT} & \multicolumn{1}{c @{\hspace{0.125\tabcolsep}}}{\bfseries $\uparrow$ BLEU} \\
\toprule
         \parbox[t]{2mm}{\multirow{11}{*}{\rotatebox[origin=c]{90}{Not Fine-Tuned}}} &

                \textsc{Cascade Unidirect} & 37.3 & 22.2 & 45.6 &  21.9 & 28.0 & 15.3 & 40.0 &  24.4 \\
&          \textsc{Cascade Unidirect Ora.} & 36.3 & 21.5 & 45.0 &  22.1 & 23.5 & 12.0 & 38.0 &  25.6 \\
&                     \textsc{E2E Unidirect} & 36.9 & 22.0 & 45.1 &  21.8 & 28.2 & 15.6 & 39.7 &  24.7 \\
&              \textsc{E2E Unidirect Ora.} & 35.7 & 21.3 & 44.4 &  22.2 & 23.2 & 12.0 & 37.6 &  26.0 \\
&           \textsc{Cascade Uni Shared Enc} & 36.0 & 20.5 & 44.7 &  21.9 & 25.6 & 13.0 & 39.8 &  24.3 \\
&     \textsc{Cascade Uni Shared Enc Ora.} & 36.0 & 20.5 & 44.2 &  22.2 & 25.6 & 13.0 & 38.8 &  24.8 \\
&              \textsc{E2E Bidirect by Lang} & 36.9 & 23.6 & 43.0 &  23.2 & 27.2 & 15.5 & 39.2 &  25.1 \\
&       \textsc{E2E Bidirect by Lang Ora.} & 36.1 & 22.9 & 42.6 &  23.3 & 25.3 & 14.0 & 38.4 &  25.5 \\
&              \textsc{E2E Bidirect by Task} & 34.1 & 19.4 & 42.3 &  23.0 & 23.6 & 11.9 & 37.4 &  26.0 \\
&                 \textsc{ Cascade Bidirect} & 37.2 & 21.3 & 43.8 &  21.8 & 26.5 & 13.3 & 39.5 &  24.1 \\
&               \textsc{E2E Bidirect Shared} & 33.8 & 19.3 & 41.5 &  23.3 & 23.2 & 11.8 & 37.1 &  26.2 \\

      \midrule
               \parbox[t]{2mm}{\multirow{11}{*}{\rotatebox[origin=c]{90}{Fine-Tuned}}} &
                \textsc{Cascade Unidirect} & 33.5 & 18.5 & 39.6 &  24.6 & 24.8 & 12.9 & 38.2 &  25.5 \\
 &         \textsc{Cascade Unidirect Ora.} & 33.1 & 18.4 & 39.4 &  24.6 & 23.8 & 12.1 & 37.7 &  25.7 \\
 &                    \textsc{E2E Unidirect} & 33.4 & 19.1 & 40.0 &  24.4 & 25.3 & 13.3 & 38.3 &  25.5 \\
 &             \textsc{E2E Unidirect Ora.} & 33.2 & 19.0 & 39.9 &  24.5 & 23.9 & 12.2 & 37.7 &  25.9 \\
 &          \textsc{Cascade Uni Shared Enc} & 31.2 & 17.1 & 38.4 &  25.4 & 25.6 & 13.0 & 38.7 &  24.8 \\
 &    \textsc{Cascade Uni Shared Enc Ora.} & 31.3 & 17.1 & 38.2 &  25.6 & 25.3 & 12.8 & 38.5 &  24.9 \\
 &             \textsc{E2E Bidirect by Lang} & 36.7 & 23.3 & 42.9 &  22.8 & 27.3 & 15.5 & 39.3 &  25.0 \\
 &      \textsc{E2E Bidirect by Lang Ora.} & 35.9 & 22.7 & 42.5 &  23.0 & 25.3 & 14.0 & 38.4 &  25.4 \\
 &             \textsc{E2E Bidirect by Task} & 30.1 & 16.2 & 38.3 &  25.6 & 24.3 & 12.3 & 37.8 &  25.6 \\
 &                 \textsc{Cascade Bidirect} & 33.2 & 18.3 & 41.0 &  23.2 & 28.1 & 14.3 & 40.1 &  23.2 \\
 &              \textsc{E2E Bidirect Shared} & 30.0 & 16.4 & 38.0 &  25.4 & 24.1 & 12.2 & 37.3 &  26.1 \\

\bottomrule
\end{tabular}
\caption{Scores on the Fisher dataset. CCT stands for Charcut. Results from zero CS training are on the top half and results after fine-tuning are on the bottom half. Ora stands for Oracle.\label{tab:fisher_all}}
\end{table*}

\end{document}